\documentclass[runningheads]{llncs}
\usepackage{xcolor}
\usepackage{amsmath}
\usepackage{graphicx}
\usepackage{amssymb}
\usepackage{multirow}
\usepackage{caption}
\usepackage{subcaption}
\usepackage{array}
\usepackage{listings}

\def\doi#1{\href{https://doi.org/\detokenize{#1}}{\url{https://doi.org/\detokenize{#1}}}}
\usepackage{graphicx}

\usepackage{hyperref}

\newcolumntype{!}{>{\global\let\currentrowstyle\relax}}

\newcolumntype{^}{>{\currentrowstyle}}

\definecolor{gray}{rgb}{0.4,0.4,0.4}
\definecolor{darkblue}{rgb}{0.0,0.0,0.6}
\definecolor{cyan}{rgb}{0.0,0.6,0.6}

\lstdefinelanguage{XML}
{
    morestring=[b]",
    morestring=[s]{>}{<},
    morecomment=[s]{<?}{?>},
    stringstyle=\color{black},
    identifierstyle=\color{darkblue},
    keywordstyle=\color{cyan},
    morekeywords={xmlns,version,type}
}

\begin{document}
\title{LDD: A Dataset for Grape Diseases Object Detection and Instance Segmentation}
\titlerunning{LDD: A Grape Diseases Dataset}
%
\author{Leonardo Rossi*\inst{1}\orcidID{0000-0002-9316-595X} \and
 Marco Valenti*\inst{1} \and
 Sara Elisabetta Legler\inst{2} \and
 Andrea Prati\inst{1}\orcidID{0000-0002-1211-529X}}
\authorrunning{L. Rossi et al.}
%
\institute{IMP Lab - D.I.A. - University of Parma - Parma, Italy\\
\email{\{leonardo.rossi, andrea.prati\}@unipr.it, marco.valenti@studenti.unipr.it}\\
\url{implab.ce.unipr.it/}\\
Horta s.r.l. - Università Cattolica del Sacro Cuore spin off in Piacenza, Italy\\
    \email{\{s.legler@horta-srl.com}\\
    \url{www.horta-srl.it/}}
%
%
\maketitle              
\begin{abstract}
The Instance Segmentation task, an extension of the well-known Object Detection task, is of great help in many areas, such as precision agriculture: being able to automatically identify plant organs and the possible diseases associated with them, allows to effectively scale and automate crop monitoring and its diseases control.
To address the problem related to early disease detection and diagnosis on vines plants, a new dataset has been created with the goal of advancing the state-of-the-art of diseases recognition via instance segmentation approaches.
This was achieved by gathering images of leaves and clusters of grapes affected by diseases in their natural context.
The dataset contains photos of 10 object types which include leaves and grapes with and without symptoms of the eight more common grape diseases, with a total of 17,706 labeled instances in 1,092 images.
Multiple statistical measures are proposed in order to offer a complete view on the characteristics of the dataset.
Preliminary results for the object detection and instance segmentation tasks reached by the models Mask R-CNN \cite{he2017mask} and $R^3$-CNN \cite{10.1007/978-3-030-89128-2_46} are provided as baseline, demonstrating that the procedure is able to reach promising results about the objective of automatic diseases' symptoms recognition.


\let\thefootnote\relax\footnotetext{* These authors contributed equally}

\end{abstract}
\section{Introduction}
Recognizing plant diseases is a deeply-felt problem in the agricultural sector.
Early disease detection and diagnosis is a key aspect for a correct and sustainable disease control.
Traditionally, the diagnosis of leaf and cluster diseases of grapes relies on expert judgment based on visual inspection of the disease symptoms and signs, which usually leads to high cost and potential errors.
With the rapid development of artificial intelligence, machine learning methods have been applied to plant disease detection to make it smarter.
In recent years, deep learning approaches led to a huge improvement of image analysis with the tasks of object detection \cite{amit2020object} and its evolution, instance segmentation \cite{hafiz2020survey}.
A main problem related with these deep learning tasks is the need of a labeled dataset with a large amount of images in order to obtain an acceptable performance.
The purpose behind the creation of this dataset is the implementation of an automatic system for the analysis of the leaves and bunches of vines through images, in order to identify the diseases that affect them.
The project was born in collaboration with Horta s.r.l, a spin off company of the Università Cattolica del Sacro Cuore in Piacenza (Italy), that develops and provides web based services to agricultural and agro-industrial chains, with the aim of increasing competitiveness and sustainability of crop management and ensuring greater food safety.
The final purpose is to give the possibility to an inexperienced user to analyze the general status of the plants and identify any diseases through a photo taken from a mobile device.
Our goal is to be able to identify each disease in a very precise way, and provide the user with information about the disease severity.
Plant diseases cause significant losses to farmers and the possibility of an early and efficient disease symptoms detection is of fundamental importance for a correct disease management.
The information provided could be integrated automatically in broader management tools, such as Decision Support Systems (DSSs), and increase the support given to farmers for the sustainable management of vineyards.
Instance segmentation represents a good tool since it allows to examine the leaves and grapes accurately at segmentation level.
\section{Related Works}
Within the literature, there are several attempts to automatically analyze plants and their diseases by means of images.
In \cite{alessandrini2021grapevine}, authors created a dataset of grapevine leaves in order to detect the Esca disease through an image classification algorithm. With the same scope, authors of \cite{ranccon2019comparison} addressed object detection for Esca disease, by providing 6,000 labelled images of Bordeaux vineyards.
In \cite{xie2020deep}, the authors proposed a dataset for three common grape leaf diseases (Black rot, Black measles, also called Esca, and Leaf blight) and for the Mites of grape;
the dataset contains 4,449 original images of grape leaf diseases labeled with bounding boxes.
Similarly, \cite{santos2020grape} gathered 300 images for five grape varieties: Chardonnay, Cabernet Franc,	Cabernet Sauvignon, Sauvignon Blanc and Syrah; the corresponding dataset is publicly available and contains label for grape detection and instance segmentation.
Another example of dataset on disease image recognition is the one proposed in \cite{wang2012application}, where labels to classify four diseases of wheat (stripe rust, leaf rust, grape downy mildew, grape powdery mildew) are also provided. Finally, in \cite{fuentes2017robust}, authors labeled with bounding boxes 5,000 images of tomato diseases.
Recently, a dataset for Apple Orchards diseases has been proposed in \cite{storey2022leaf}, which contains handmade annotated segmentation of a image subset (142) from Plant Pathology Challenge 2020 database \cite{thapa2020plant}.
The \cite{udawant2022cotton} is an example of a dataset for instance segmentation for cotton leaf disease detection, with 2,000 photos made with a smartphone camera in a real-world scenario.

All the datasets that take grape diseases into consideration are limited to image classification or, at most, object detection.
On the other hand, the datasets that offer segmentation consider grape varieties or diseases of other plant.
As far as we know, no one has ever composed a dataset containing bounding boxes and segmentations for grape diseases before.
\section{Leaf Diseases Dataset}
\begin{figure}[h]
    \centering
    \begin{subfigure}{0.49\linewidth}
        \includegraphics[width=.9\linewidth]{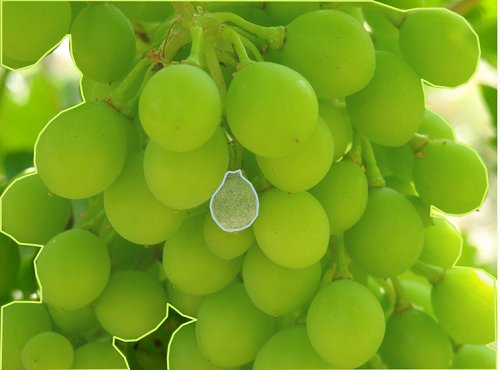}
        \caption{bunch of grapes affected by powdery mildew (oidio).}
        \label{fig:ldd-img1}
    \end{subfigure}
    \begin{subfigure}{0.49\linewidth}
        \includegraphics[width=.9\linewidth]{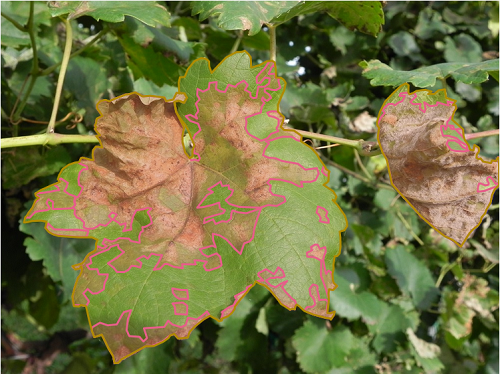}
        \caption{leaves affected by downy mildew (peronospora).}
        \label{fig:ldd-img2}
    \end{subfigure}
 \caption{Example of segmentation with the help of Label Studio.}
\label{fig:ldd-img_labeled}
\end{figure}

Our dataset contains 1,092 RGB images of grapes and 17,706 annotations in COCO-like \cite{lin2014microsoft} format for the tasks of Object Detection and Instance Segmentation.
The images were collected from Horta's internal databases, the competition \textit{Grapevine Disease Images} \footnote{https://www.kaggle.com/piyushmishra1999/plantvillage-grape}, and from search engine scraping;
all segmentations were manually made by the authors.
Figure \ref{fig:ldd-img_labeled} shows some examples of labeled images.
The annotations are based on 10 categories of objects: leaf, grape and eight different diseases.
Table \ref{tab:ldd-dataset-count} shows the the class id, label key and name, the type of disease (together with leaf and grape), and the total available annotations (column "Count").

\begin{table}
    \centering
\begin{tabular}{|c|r|r|l|l|r|}
    \hline
     & Key & Count & Label & Name & Type \\
    \hline
    \hline
     1 & LBR & 5478 & vl\_black\_rot & Black Rot & leaf disease \\
     2 & GBR & 1443 & vg\_black\_rot & Black Rot & grape disease \\
     3 & LGM &   88 & vl\_grey\_mould & Grey mould (Botrite) & leaf disease \\
     4 & GGM &  612 & vg\_grey\_mould & Grey mould (Botrite) & grape disease \\
     6 &  VL & 1647 & vines\_leaf & Vine leaf & leaf \\
     7 &  VG &  993 & vines\_grape & Bunch of grapevines & grape \\
     9 & LPM & 2086 & vl\_powdery\_mildew & Powdery mildew (Oidio) & leaf disease \\
    10 & GPM & 1740 & vg\_powdery\_mildew & Powdery mildew (Oidio) & grape disease \\
    12 & LDM & 2565 & vl\_downy\_mildew & Downy mildew (Peronospora) & leaf disease \\
    13 & GDM & 1054 & vg\_downy\_mildew & Downy mildew (Peronospora) & grape disease \\
    \hline
       &     & 17706 & \multicolumn{3}{l|}{Total} \\
    \hline
\end{tabular}
\caption{LDD annotations description.}
\label{tab:ldd-dataset-count}
\end{table}



Annotations are split in training (80\%) and validation (20\%) dataset, with the help of the \textit{cocosplit} \footnote{https://github.com/akarazniewicz/cocosplit} utility.

\begin{figure}[h]
    \begin{center}
        \includegraphics[trim=0 0 0 0, clip, width=1.\linewidth]{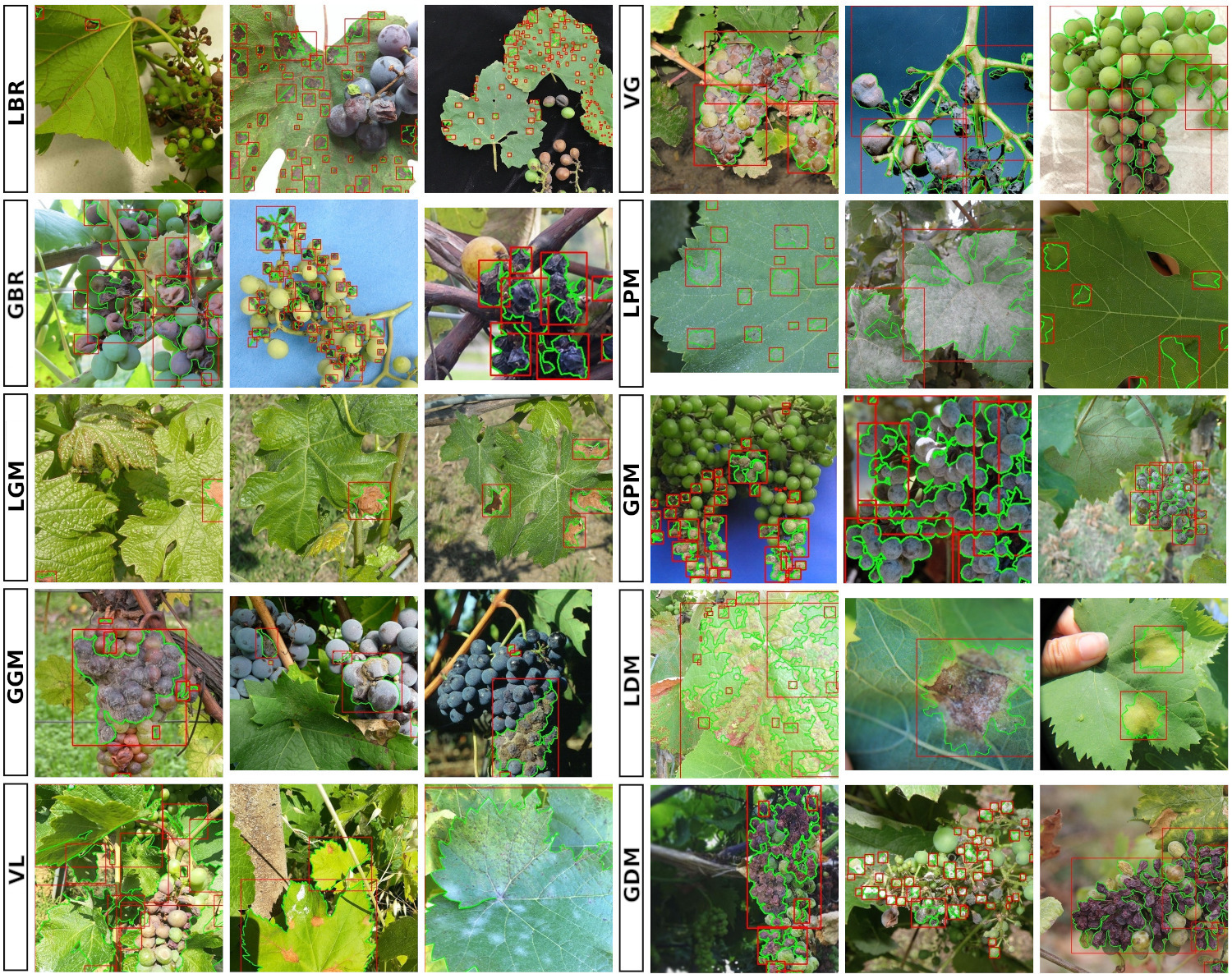}
    \end{center}
    \caption{Samples of annotated images in LDD.}
    \label{fig:ldd-collage}
\end{figure}

Figure \ref{fig:ldd-collage} reports some examples for each class.

\subsubsection{Annotation procedure.}

To insert the annotations, the online tool \textit{Label Studio v1.5} \footnote{https://labelstud.io/} was used with the following configuration:

\begin{lstlisting}[language=XML, caption="Label Studio Instance Segmentation configuration", basicstyle=\footnotesize]
<View>
    <Image name="image" value="$image" zoom="true" zoomControl="true"/>
    <PolygonLabels name="label" toName="image" strokeWidth="3"
            pointSize="small" opacity="0.9">
        <Label value="vl_black_rot" background="#D4380D"/>
        <Label value="vg_black_rot" background="#FFC069"/>
        <Label value="vl_grey_mould" background="#AD8B00"/>
        <Label value="vg_grey_mould" background="#D3F261"/>
        <Label value="vl_powdery_mildew" background="#096DD9"/>
        <Label value="vg_powdery_mildew" background="#ADC6FF"/>
        <Label value="vl_downy_mildew" background="#F759AB"/>
        <Label value="vg_downy_mildew" background="#FFA39E"/>
        <Label value="vines_leaf" background="#AD8B00"/>
        <Label value="vines_grape" background="#D3F261"/>
    </PolygonLabels>
</View>
\end{lstlisting}

\begin{figure}[h]
    \begin{center}
        \includegraphics[trim=0 0 0 0, clip, width=.4\linewidth]{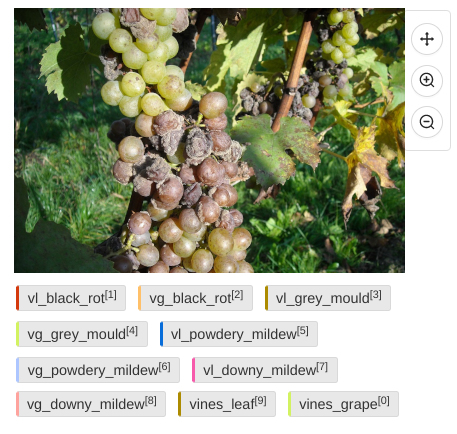}
    \end{center}
    \caption{Label Studio web interface.}
    \label{fig:ldd-labelstudio}
\end{figure}

Example of web interface can be seen in Figure \ref{fig:ldd-labelstudio} where, for each image, the user can select the proper category and then the polygon which identifies the instance.

Since the annotation of this type of images also requires expertise in agronomy, we first had a student creating coarse-grain annotations, and then have an agronomist expert in this field to review and fix them.
At the end, the annotations have been exported in COCO-like format, using the built-in export function of label studio that for each polygonal mask auto-generate the corresponding bounding box.

\subsubsection{Statistics.}

\begin{figure}[h]
    \centering
    \begin{subfigure}[t]{0.59\linewidth}
        \includegraphics[width=1.\linewidth, height=14em]{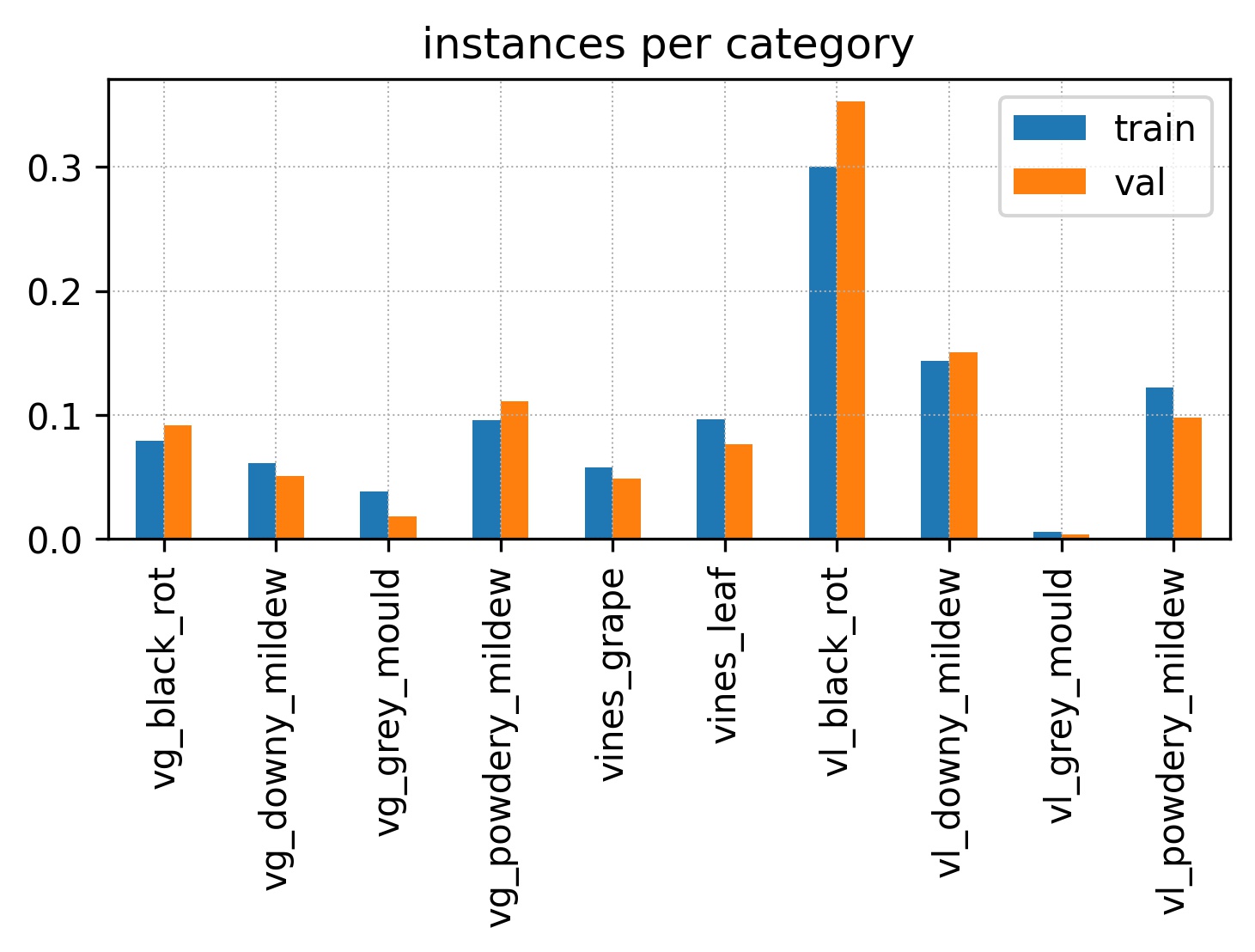}
        \caption{Percentage of instances per category.}
        \label{fig:ldd-instances-per-cats}
    \end{subfigure}
    \begin{subfigure}[t]{0.39\linewidth}
        \includegraphics[width=1.\linewidth, height=14em]{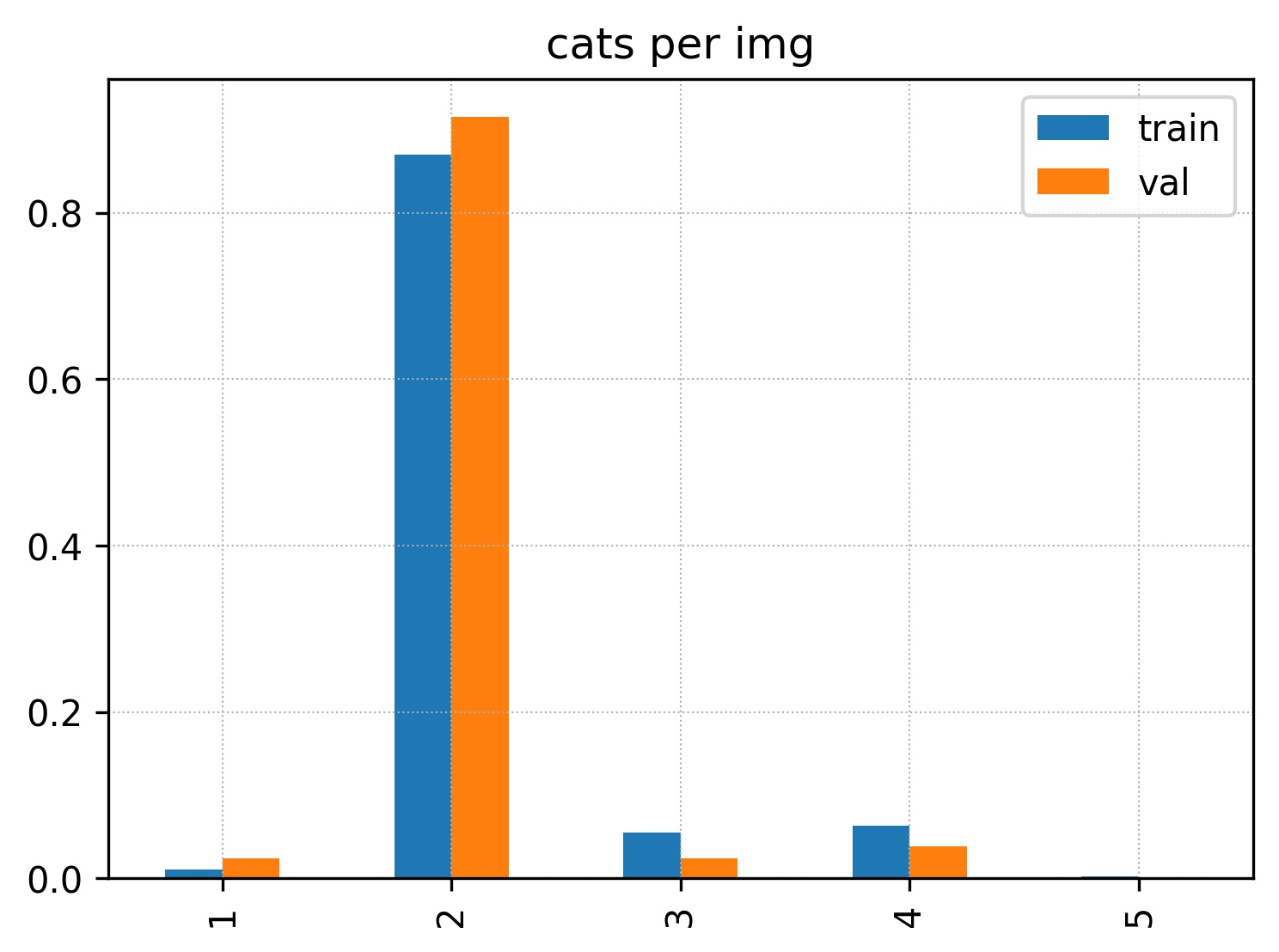}
        \caption{Categories per image}
        \label{fig:ldd-cats-per-img}
    \end{subfigure}
    \caption{Dataset LDD:
        (\ref{fig:ldd-instances-per-cats}) Percentage of instances for each category.
        (\ref{fig:ldd-cats-per-img}) Percentage of annotated categories per image.
        Better seen in color.}
    \label{fig:ldd-stats-categories}
\end{figure}

\begin{figure}[h]
    \begin{center}
        \includegraphics[width=1.\linewidth]{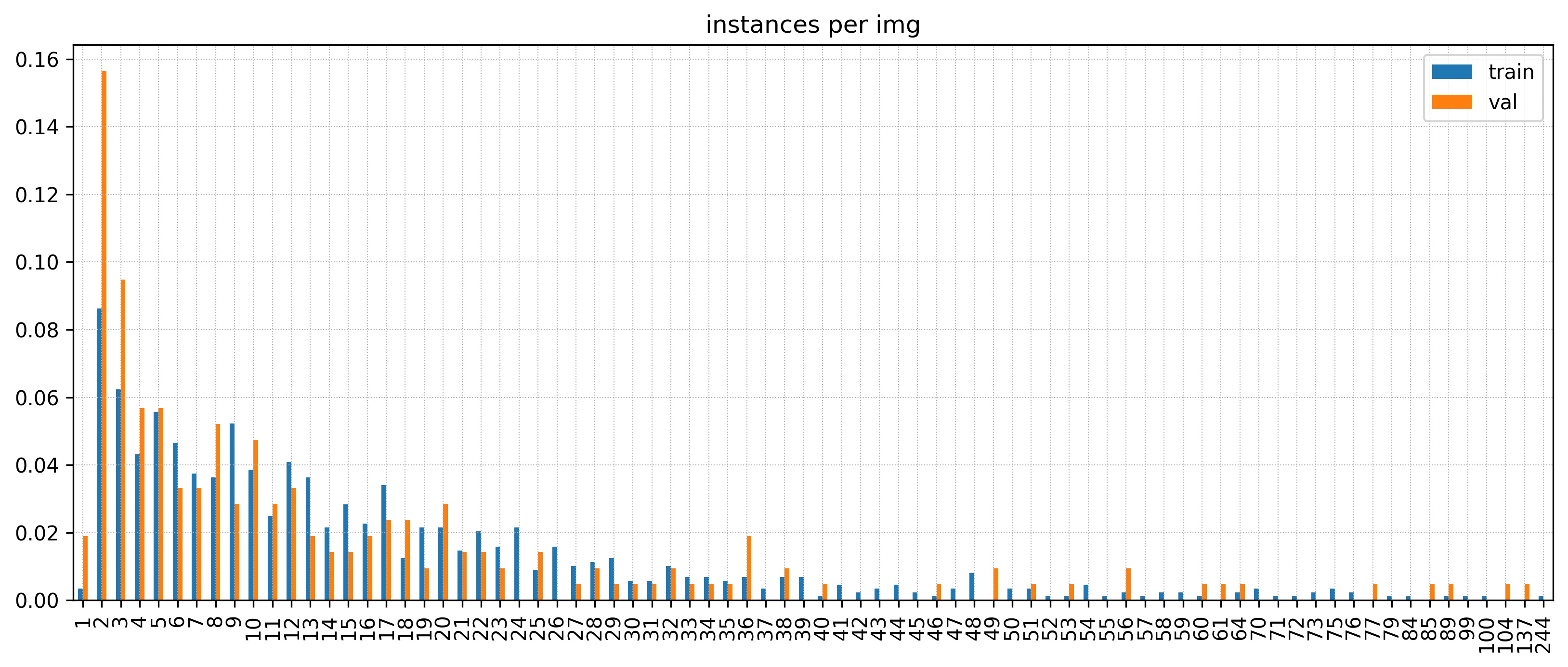}
    \end{center}
    \caption{Annotations per image: percentage of images in LDD.}
    \label{fig:ldd-instances-per-img}
\end{figure}

\begin{figure}[h]
    \centering

    \begin{subfigure}{0.49\linewidth}
        \includegraphics[width=1.\linewidth]{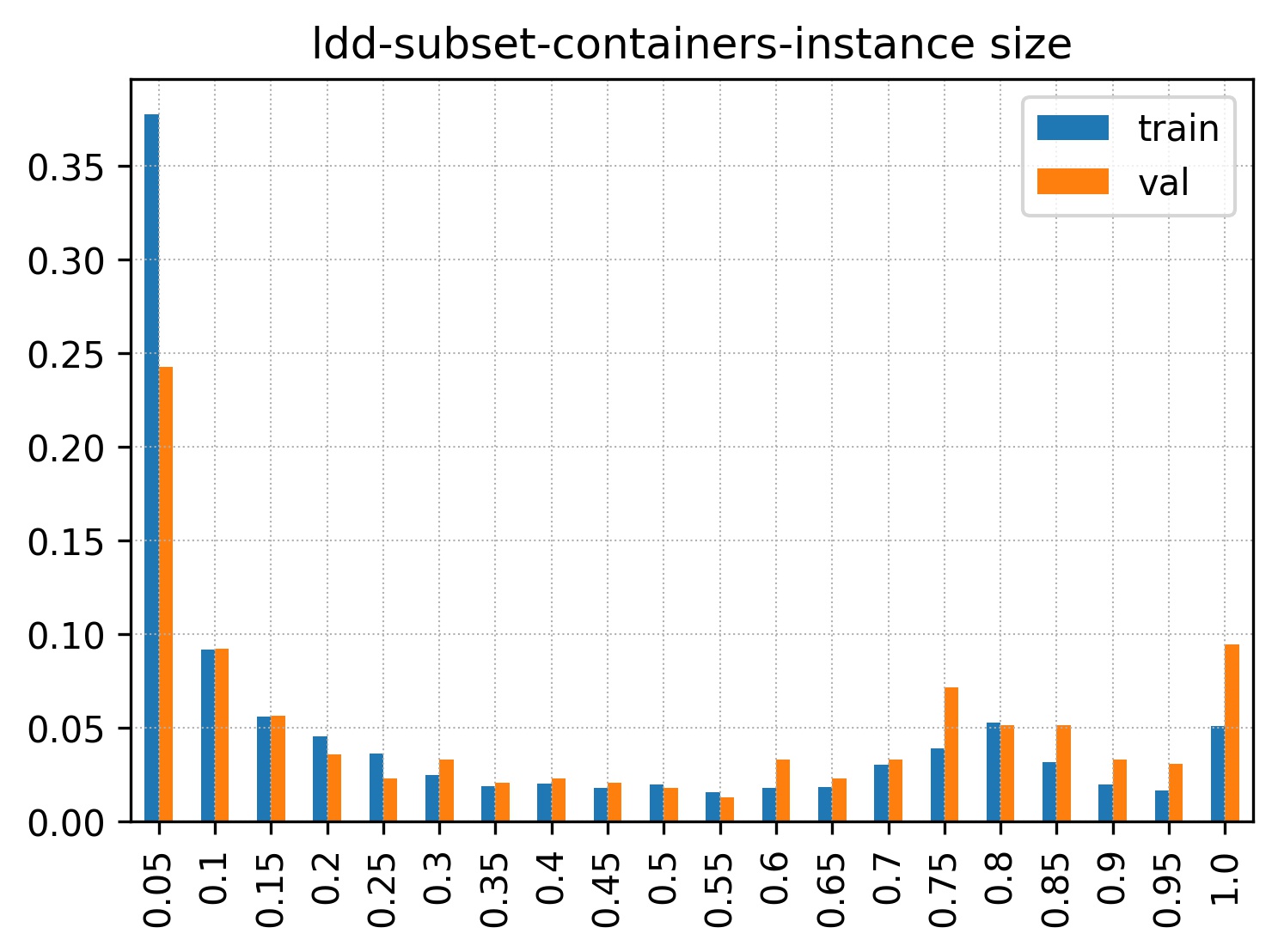}
        \caption{leaves and grapes instances}
        \label{fig:ldd-instance-size-contaieners}
    \end{subfigure}
        \begin{subfigure}{0.49\linewidth}
        \includegraphics[width=1.\linewidth]{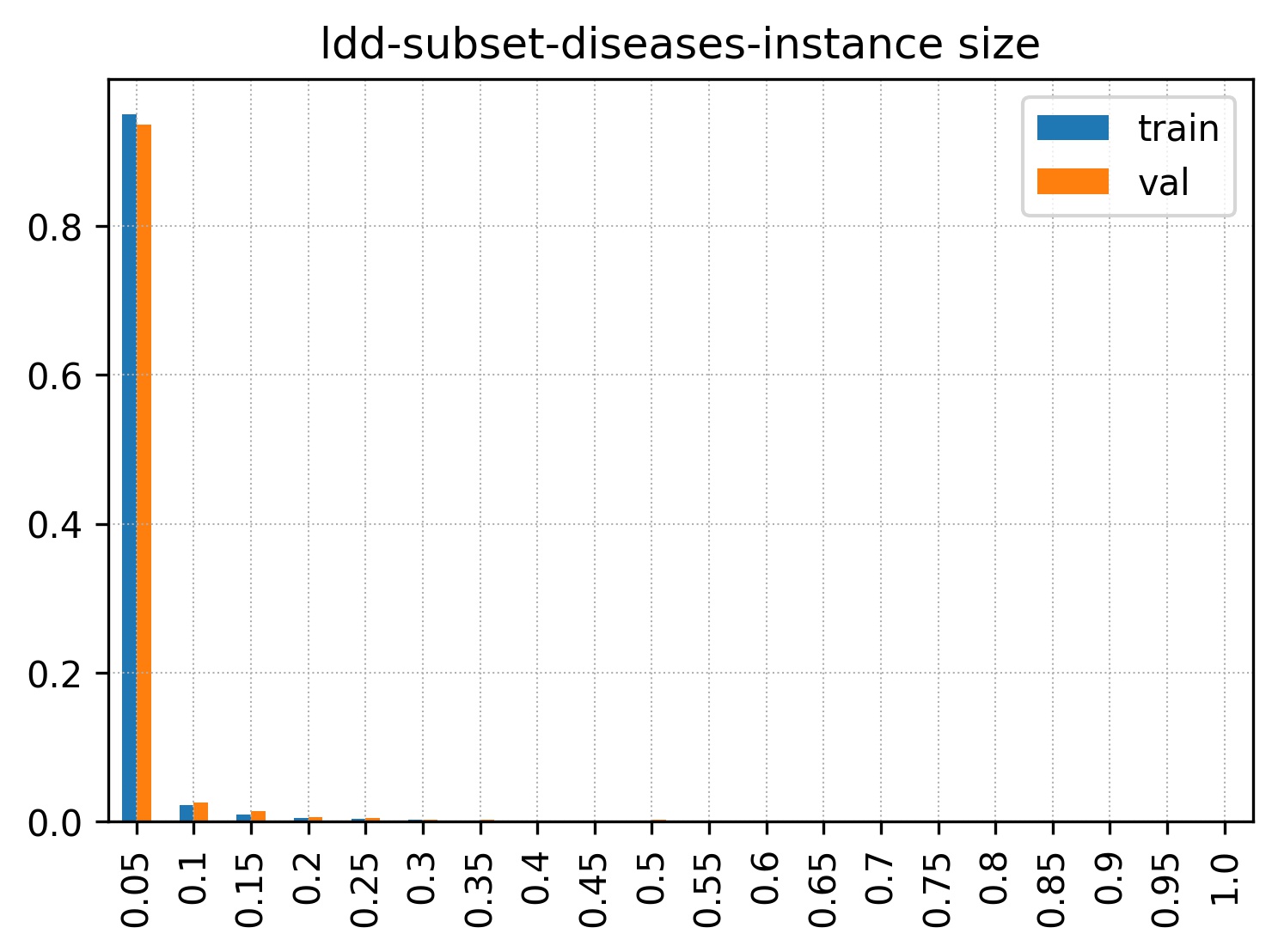}
        \caption{diseases instances}
        \label{fig:ldd-instance-size-diseases}
    \end{subfigure}
    \begin{subfigure}{0.49\linewidth}
        \includegraphics[width=1.\linewidth]{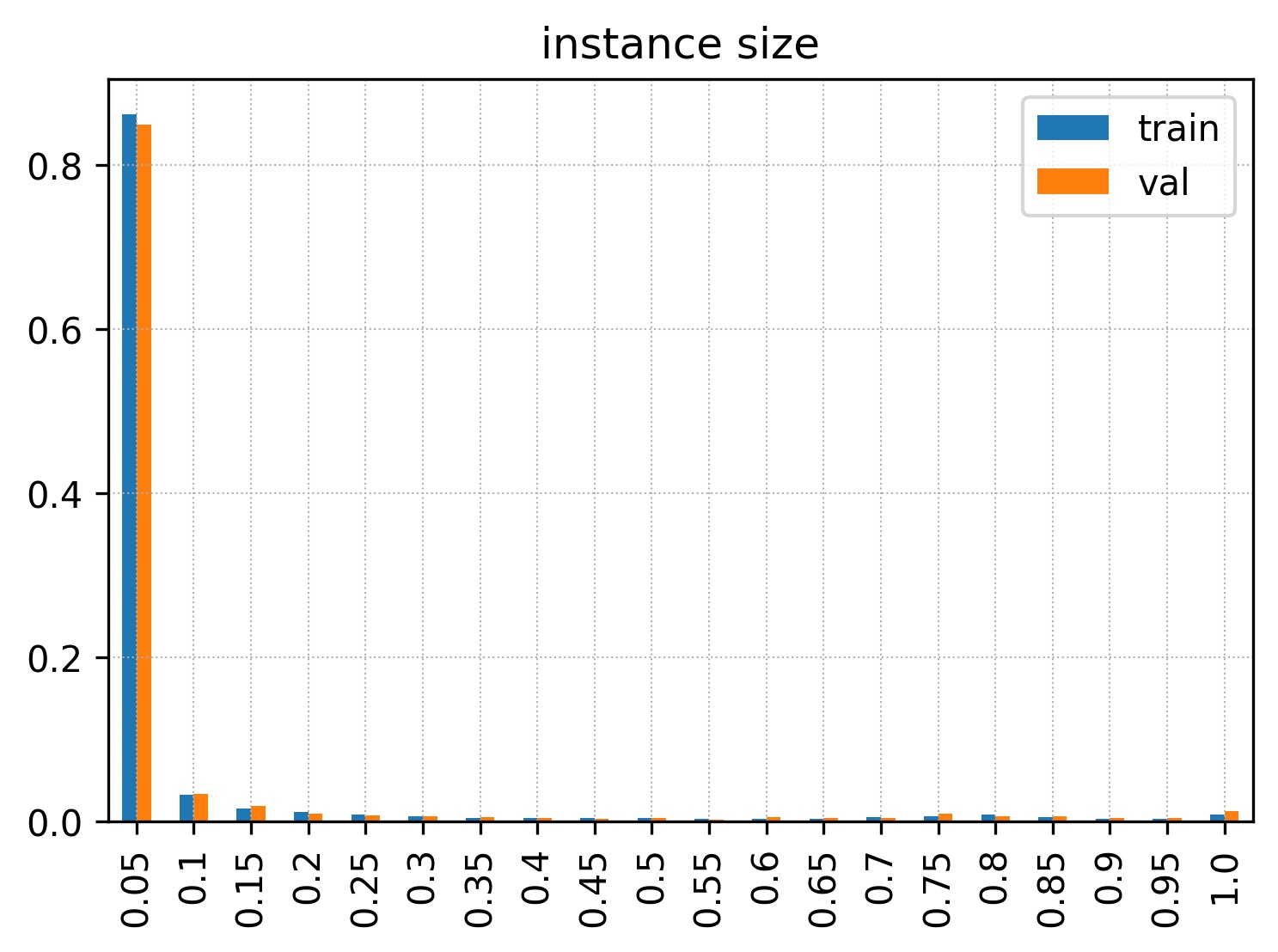}
        \caption{all instances}
        \label{fig:ldd-instance-size-all}
    \end{subfigure}
    \caption{Dataset LDD instance sizes distribution:
        (\ref{fig:ldd-instance-size-contaieners}) only leaves and Grapes.
        (\ref{fig:ldd-instance-size-diseases}) only diseases.
        (\ref{fig:ldd-instance-size-all}) all dataset.
        Better seen in color.}
    \label{fig:ldd-instance-size}
\end{figure}

In Figure \ref{fig:ldd-instances-per-cats}, the percentage of instances for each category are reported.
As it can be noticed, the dataset is strongly unbalanced, but the same distribution can be appreciated in both training and validation dataset.
As shown in Figure \ref{fig:ldd-cats-per-img}, the maximum number of unique categories for each image can vary from one to five, with more than 80\% of the images containing two classes.
This could be explained by the fact that, usually, the image contains a leaf or a grape affected by a particular disease.
Contrary to the number of classes per image, the number of annotations within a single image is much more varied, as shown in Figure \ref{fig:ldd-instances-per-img}, with more than one hundred different annotations for complex images.
Figures \ref{fig:ldd-instance-size} offer a different point of view, showing that the mean percentage of the instance size with respect to the image size is very low, with more than 80\% of the instances being in the range $[0\%, 0.05\%]$ (see, for instance Fig. \ref{fig:ldd-instance-size-diseases} and \ref{fig:ldd-instance-size-all}).
Figure \ref{fig:ldd-instance-size-contaieners} shows the distributions for leaves and grapes, which are much more diverse with respect to diseases.

\section{Experiments}
\noindent\textbf{Evaluation Metrics}.
All the tests are done on LDD validation dataset, which contains 212 images and 3,234 annotations.
We report mean Average Precision (mAP) and $AP_{50}$ and $AP_{75}$ with 0.5 and 0.75 minimum IoU thresholds, and $AP_s$, $AP_m$ and $AP_l$ for small, medium and large objects, respectively.

\noindent\textbf{Implementation details.}
All the values are obtained running the training with the same hardware and hyper-parameters.
When available, the original code released by the authors is used.
Our code is developed on top of the MMDetection \cite{mmdetection} source code.
We perform the training on a single machine with 1 NVIDIA GeForce RTX 2070 with 8GB of memory.
The train lasts 12 epochs with a batch size of 2 images.
We use the Stochastic Gradient Descent (SGD) optimization algorithm with a learning rate of 0.00125, a weight decay of 0.0001, and a momentum of 0.9.
The learning rate decays at epoch 8 and 11.
We use the ResNet 50 \cite{he2016deep} backbone for the models, initialized by pre-trained weights on ImageNet.

For the experiment, basic Mask R-CNN \cite{he2017mask} and $R^3$-CNN \cite{10.1007/978-3-030-89128-2_46} models were used, with the intention to test these baseline models on the proposed dataset.

\begin{table*}[bth]
    \centering
    \footnotesize
    \setlength{\tabcolsep}{1.2pt}
    \begin{tabular}{@{}c|l||c|c|c|c|c|c||c|c|c|c|c|c}
        & & \multicolumn{6}{c||}{Bounding Box} & \multicolumn{6}{c}{Mask} \\
\# & Method & $AP$ & $AP_{50}$ & $AP_{75}$  & $AP_{s}$ & $AP_{m}$ & $AP_{l}$  & AP & $AP_{50}$ & $AP_{75}$  & $AP_{s}$ & $AP_{m}$ & $AP_{l}$\\
\hline\hline
1 & Mask R-CNN  & 21.0 & 36.8 & 21.9 & 9.1 & 18.8 & 21.2 & 20.2 & 35.6 & 21.6 & 8.8 & 17.2 & 20.1 \\
2 & $R^3$-CNN   & 22.7 & 38.4 & 22.8 & 9.5 & 19.9 & 30.9 & 22.2 & 36.2 & 24.3 & 9.4 & 19.3 & 29.8 \\
    \end{tabular}
    \caption{Performance of Mask R-CNN and $R^3$-CNN models.}
    \label{tab:ldd-nets-results}
\end{table*}

\begin{figure*}[bth]
    \begin{center}
        \includegraphics[trim=0 0 0 0, clip, width=1.\linewidth]{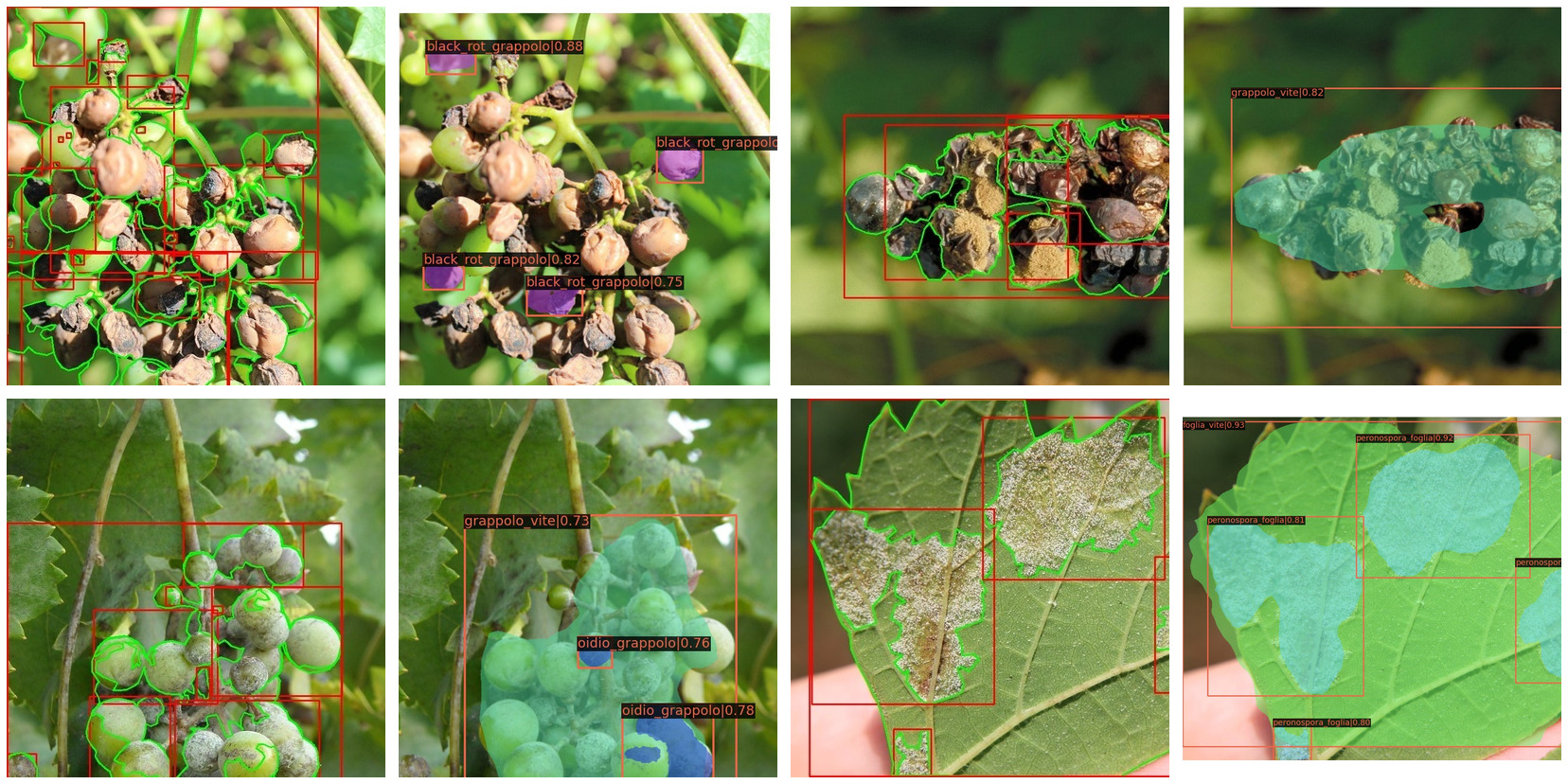}
    \end{center}
    \caption{$R^3$-CNN prediction examples (right) compared with ground-truth (left).}
    \label{fig:ldd-r3cnn-preds}
\end{figure*}

\begin{figure*}[bth]
    \begin{center}
        \includegraphics[width=.8\linewidth]{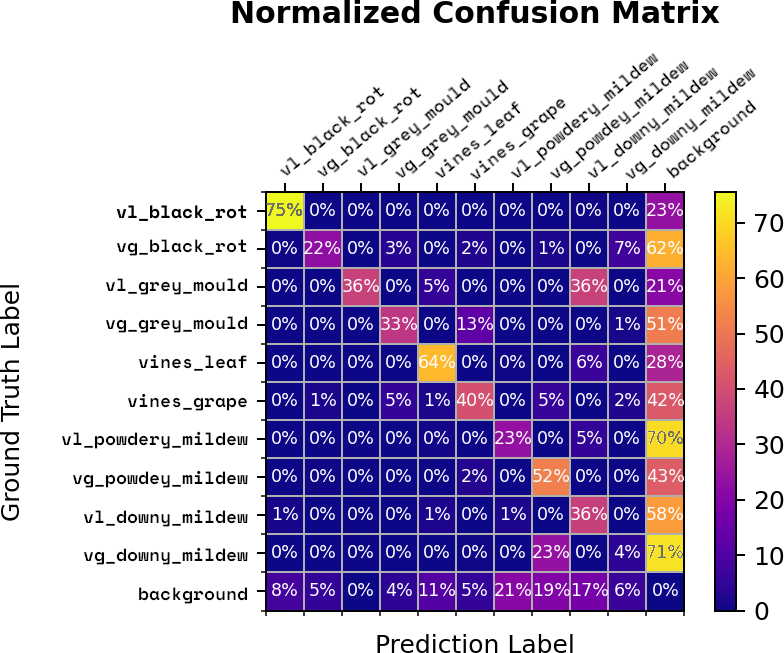}
    \end{center}
    \caption{$R^3$-CNN confusion matrix.}
    \label{fig:ldd-r3cnn-cm}
\end{figure*}

Table \ref{tab:ldd-nets-results} shows the training results for the task of Object Detection and Instance Segmentation.
Best values are given by the $R^3$-CNN.
More in detail, Tables \ref{tab:ldd-bbox-per-class} and \ref{tab:ldd-segm-per-class} show AP values for each category for object detection and instance segmentation, respectively.
Figure \ref{fig:ldd-r3cnn-preds} shows some examples of $R^3$-CNN predictions compared with ground-truth.
For black rot on leaves, one of most represented disease in our dataset, we reached promising results.
The same for leaves, even though there were much fewer instances.
Conversely, a grape of wine is more difficult to detect achieving a third of performances compared to leaves with two thirds of its instances.
Despite a fair number of instances, most difficult disease to detect is \textit{peronospora\_grappolo} (Downy mildew on bunches).
Figure \ref{fig:ldd-r3cnn-cm} shows the confusion matrix for the $R^3$-CNN model.

\begin{table*}[bth]
    \centering
    \small
    \setlength{\tabcolsep}{1.2pt}
    \begin{tabular}{@{}c|l||c|c|c|c|c|c|c|c|c|c|c}
        \# & Method & LBR & GBR & LGM & GGM & VL & VG & LPM & GPM & LDM & GDM & AP \\
        \hline\hline
        1 & Mask R-CNN  & 45.8 & 14.3 & 41.7 & 6.0 & 51.3 & 14.0 & 4.8 & 16.9 & 14.2 & 0.9 & 21.0 \\
        2 & $R^3$-CNN   & 49.1 & 15.2 & 40.4 & 7.0 & 55.3 & 16.7 & 6.0 & 18.2 & 17.4 & 1.4 & 22.7 \\
    \end{tabular}
\caption{BBox AP per category. See Table \ref{tab:ldd-dataset-count} for class key. AP: Average Precision.}
\label{tab:ldd-bbox-per-class}
\end{table*}

\begin{table*}[t!]
    \centering
    \small
    \setlength{\tabcolsep}{1.2pt}
    \begin{tabular}{@{}c|l||c|c|c|c|c|c|c|c|c|c|c}
        \# & Method & LBR & GBR & LGM & GGM & VL & VG & LPM & GPM & LDM & GDM & AP \\
        \hline\hline
        1 & Mask R-CNN  & 48.6 & 15.0 & 40.4 & 5.4 & 50.3 & 10.3 & 4.4 & 14.3 & 12.1 & 0.7 & 20.2 \\
        2 & $R^3$-CNN   & 52.0 & 16.2 & 44.6 & 4.6 & 54.8 & 11.6 & 5.0 & 15.8 & 15.7 & 1.3 & 22.2 \\
    \end{tabular}
    \caption{Segmentation AP per category. See Table \ref{tab:ldd-dataset-count} for class key. AP: Average Precision.}
    \label{tab:ldd-segm-per-class}
\end{table*}

\section{Discussion on the Dataset}
We introduced a new dataset for segmenting the most common leaf and grapes diseases in their natural environment.
The main idea behind the creation of this dataset was to use images with few leaves or bunches affected by a few diseases in foreground in order to facilitate and speed-up the manual process of labeling, but obtaining an accurate annotation.
This choice explains the number of categories per image shown in Figure \ref{fig:ldd-cats-per-img} and can possibly led to a degradation of performances in case of the segmentation of images with an high numbers of leaves or grapes, that are very common into a natural environment.

Emphasis was placed on generating ground truth labels where the disease's polygon was entirely contained inside the leaf or cluster polygon, in order to encourage the strict connection between infected organ and disease, trying to avoid cases where an isolated disease is shown.

Another peculiarity of this dataset is related to the choice of decomposing large ground-truth polygons into smaller ones in order to realize the most accurate labeling.
In particular, this approach was used for the labeling of clusters and their diseases in order to exclude the small gaps that are present between berries in a bunch.
A consequence of this label partitioning is the predominance of small instances in the dataset, as shown in Figure \ref{fig:ldd-instance-size-all}, which can led to a more challenging segmentation task.
\section{Conclusions}
In this paper, we have described the newly-proposed LDD dataset, which contains images coming mostly from the natural environment in which the vines are cultivated. Images are cataloged in terms of objects found in the images, by means of bounding boxes and segmentations in the COCO-like format.

As future work, we plan to increase the amount of images, in particular for those categories that under-represented like \textit{Powdery (Oidio)} and \textit{Downy (Peronospora)} mildews on both grapes and leaves.
The final goal is to collect at least the amount of 500 images of each disease in order to provide a common and public ground of comparison for instance segmentation architectures in this kind of applications.

\subsection*{Acknowledgments}

We acknowledge Horta s.r.l for the collaboration and help to build the dataset.
This study was performed within the project ”AGREED - Agriculture, Green \& Digital” (ARS01 00254 on PON “Ricerca e Innovazione” 2014-2020 and FSC funds).

\bibliographystyle{splncs04}
\bibliography{paper}

\begin{thebibliography}{10}
\providecommand{\url}[1]{\texttt{#1}}
\providecommand{\urlprefix}{URL }
\providecommand{\doi}[1]{https://doi.org/#1}

\bibitem{alessandrini2021grapevine}
Alessandrini, M., Rivera, R.C.F., Falaschetti, L., Pau, D., Tomaselli, V.,
  Turchetti, C.: A grapevine leaves dataset for early detection and
  classification of esca disease in vineyards through machine learning. Data in
  Brief  \textbf{35},  106809 (2021)

\bibitem{amit2020object}
Amit, Y., Felzenszwalb, P., Girshick, R.: Object detection. Computer Vision: A
  Reference Guide pp.~1--9 (2020)

\bibitem{mmdetection}
Chen, K., Wang, J., Pang, J., Cao, Y., Xiong, Y., Li, X., Sun, S., Feng, W.,
  Liu, Z., Xu, J., Zhang, Z., Cheng, D., Zhu, C., Cheng, T., Zhao, Q., Li, B.,
  Lu, X., Zhu, R., Wu, Y., Dai, J., Wang, J., Shi, J., Ouyang, W., Loy, C.C.,
  Lin, D.: Mmdetection: Open mmlab detection toolbox and benchmark. arXiv
  preprint arXiv:1906.07155  (2019)

\bibitem{fuentes2017robust}
Fuentes, A., Yoon, S., Kim, S.C., Park, D.S.: A robust deep-learning-based
  detector for real-time tomato plant diseases and pests recognition. Sensors
  \textbf{17}(9), ~2022 (2017)

\bibitem{hafiz2020survey}
Hafiz, A.M., Bhat, G.M.: A survey on instance segmentation: state of the art.
  International journal of multimedia information retrieval pp. 1--19 (2020)

\bibitem{he2017mask}
He, K., Gkioxari, G., Doll{\'a}r, P., Girshick, R.: Mask r-cnn. In: Proceedings
  of the IEEE international conference on computer vision. pp. 2961--2969
  (2017)

\bibitem{he2016deep}
He, K., Zhang, X., Ren, S., Sun, J.: Deep residual learning for image
  recognition. In: Proceedings of the IEEE conference on computer vision and
  pattern recognition. pp. 770--778 (2016)

\bibitem{lin2014microsoft}
Lin, T.Y., Maire, M., Belongie, S., Hays, J., Perona, P., Ramanan, D.,
  Doll{\'a}r, P., Zitnick, C.L.: Microsoft coco: Common objects in context. In:
  European conference on computer vision. pp. 740--755. Springer (2014)

\bibitem{ranccon2019comparison}
Ran{\c{c}}on, F., Bombrun, L., Keresztes, B., Germain, C.: Comparison of sift
  encoded and deep learning features for the classification and detection of
  esca disease in bordeaux vineyards. Remote Sensing  \textbf{11}(1), ~1 (2019)

\bibitem{10.1007/978-3-030-89128-2_46}
Rossi, L., Karimi, A., Prati, A.: Recursively refined r-cnn: Instance
  segmentation with self-roi rebalancing. In: Tsapatsoulis, N., Panayides, A.,
  Theocharides, T., Lanitis, A., Pattichis, C., Vento, M. (eds.) Computer
  Analysis of Images and Patterns. pp. 476--486. Springer International
  Publishing, Cham (2021)

\bibitem{santos2020grape}
Santos, T.T., de~Souza, L.L., dos Santos, A.A., Avila, S.: Grape detection,
  segmentation, and tracking using deep neural networks and three-dimensional
  association. Computers and Electronics in Agriculture  \textbf{170},  105247
  (2020)

\bibitem{storey2022leaf}
Storey, G., Meng, Q., Li, B.: Leaf disease segmentation and detection in apple
  orchards for precise smart spraying in sustainable agriculture.
  Sustainability  \textbf{14}(3), ~1458 (2022)

\bibitem{thapa2020plant}
Thapa, R., Zhang, K., Snavely, N., Belongie, S., Khan, A.: The plant pathology
  challenge 2020 data set to classify foliar disease of apples. Applications in
  Plant Sciences  \textbf{8}(9),  e11390 (2020)

\bibitem{udawant2022cotton}
Udawant, P., Srinath, P.: Cotton leaf disease detection using instance
  segmentation. Journal of Cases on Information Technology (JCIT)
  \textbf{24}(4),  1--10 (2022)

\bibitem{wang2012application}
Wang, H., Li, G., Ma, Z., Li, X.: Application of neural networks to image
  recognition of plant diseases. In: 2012 International Conference on Systems
  and Informatics (ICSAI2012). pp. 2159--2164. IEEE (2012)

\bibitem{xie2020deep}
Xie, X., Ma, Y., Liu, B., He, J., Li, S., Wang, H.: A deep-learning-based
  real-time detector for grape leaf diseases using improved convolutional
  neural networks. Frontiers in plant science  \textbf{11}, ~751 (2020)

\end{thebibliography}

\end{document}